\documentclass{article}

\usepackage[final]{nips_2016}

\usepackage[utf8]{inputenc} 
\usepackage[T1]{fontenc}    
\usepackage{hyperref}       
\usepackage{url}            
\usepackage{booktabs}       
\usepackage{amsfonts}       
\usepackage{nicefrac}       
\usepackage{microtype}      
\usepackage{graphicx}       
\usepackage{subcaption}     

\title{Sampling Generative Networks}

\author{
  Tom ~White\\
  School of Design\\
  Victoria University of Wellington\\
  Wellington, New Zealand \\
  \texttt{tom.white@vuw.ac.nz} \\
}

\begin{document}

\maketitle

\begin{abstract}
We introduce several techniques for sampling and visualizing the latent spaces of generative models. Replacing linear interpolation with spherical linear interpolation prevents diverging from a model's prior distribution and produces sharper samples. J-Diagrams and MINE grids are introduced as visualizations of manifolds created by analogies and nearest neighbors. We demonstrate two new techniques for deriving attribute vectors: bias-corrected vectors with data replication and synthetic vectors with data augmentation.  Binary classification using attribute vectors is presented as a technique supporting quantitative analysis of the latent space. Most techniques are intended to be independent of model type and examples are shown on both Variational Autoencoders and Generative Adversarial Networks.
\end{abstract}

\section{Introduction}
Generative models are a popular approach to unsupervised machine learning. Generative neural network models are trained to produce data samples that resemble the training set. Because the number of model parameters is significantly smaller than the training data, the models are forced to discover efficient data representations. These models are sampled from a set of latent variables in a high dimensional space, here called a latent space. Latent space can be sampled to generate observable data values. Learned latent representations often also allow semantic operations with vector space arithmetic (Figure 1).

\begin{figure}[h]
  \centering
  \includegraphics[width=7cm]{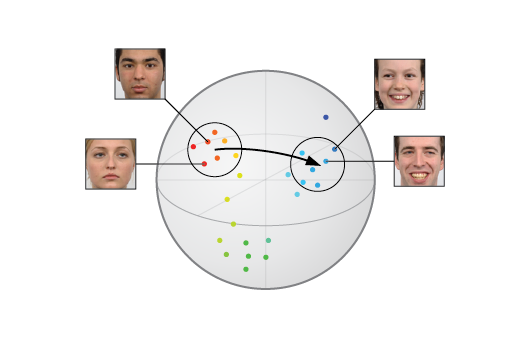}
  \caption{Schematic of the latent space of a generative model. In the general case, a generative model includes an encoder to map from the feature space (here images of faces) into a high dimensional latent space. Vector space arithmetic can be used in the latent space to perform semantic operations. The model also includes a decoder to map from the latent space back into the feature space, where the semantic operations can be observed. If the latent space transformation is the identity function we refer to the encoding and decoding as a reconstruction of the input through the model. }
\end{figure}

Generative models are often applied to datasets of images. Two popular generative models for image data are the Variational Autoencoder (VAE, Kingma \& Welling, 2014) and the Generative Adversarial Network (GAN, Goodfellow et al., 2014). VAEs use the framework of probabilistic graphical models with an objective of maximizing a lower bound on the likelihood of the data. GANs instead formalize the training process as a competition between a generative network and a separate discriminative network. Though these two frameworks are very different, both construct high dimensional latent spaces that can be sampled to generate images resembling training set data. Moreover, these latent spaces are generally highly structured and can enable complex operations on the generated images by simple vector space arithmetic in the latent space (Larsen et al., 2016).

Generative models are beginning to find their way out of academia and into creative applications. In this paper we present techniques for improving the visual quality of generative models that are generally independent of the model itself. These include spherical linear interpolation, visualizing analogies with J-diagrams, and generating local manifolds with MINE grids. These techniques can be combined generate low dimensional embeddings of images close to the trained manifold. These can be used for visualization and creating realistic interpolations across latent space. Also by standardizing these operations independent of model type, the latent space of different generative models are more directly comparable with each other, exposing the strengths and weaknesses of various approaches.

Additionally, two new techniques for building latent space attribute vectors are introduced. On labeled datasets with correlated labels, data replication can be used to create bias-corrected vectors. Synthetic attributes vectors also can be derived via data augmentation on unlabeled data. Quantitative analysis of attribute vectors can be performed by using them as the basis for attribute binary classifiers.

\section{Sampling Techniques}

Generative models are often evaluated by examining samples from the latent space. Techniques frequently used are random sampling and linear interpolation. But often these can result in sampling the latent space from locations very far outside the manifold of probable locations.

Our work has followed two useful principles when sampling the latent space of a generative model. The first is to avoid sampling from locations that are highly unlikely given the prior of the model. This technique is very well established - including being used in the original VAE paper which adjusted sampling through the inverse CDF of the Gaussian to accommodate the Gaussian prior (Kingma \& Welling, 2014). The second principle is to recognize that the dimensionality of the latent space is often artificially high and may contains dead zones that are not on the manifold learned during training. This has been demonstrated for VAE models (Makhzani et al., 2016) and implies that simply matching the model’s prior will not always be sufficient to yield samples that appear to have been drawn from the training set.

\subsection{Interpolation}

Interpolation is used to traverse between two known locations in latent space. Research on generative models often uses interpolation as a way of demonstrating that a generative model has not simply memorized the training examples (eg: Radford et al., 2015, §6.1). In creative applications interpolations can be used to provide smooth transitions between two decoded images.

Frequently linear interpolation is used, which is easily understood and implemented. But this is often inappropriate as the latent spaces of most generative models are high dimensional (> 50 dimensions) with a Gaussian or uniform prior. In such a space, linear interpolation traverses locations that are extremely unlikely given the prior.
As a concrete example, consider a 100 dimensional space with the Gaussian prior $\mu$=0, $\sigma$=1. Here all random vectors will generally a length very close to 10 (standard deviation < 1). However, linearly interpolating between any two will usually result in a "tent-pole" effect as the magnitude of the vector decreases from roughly 10 to 7 at the midpoint, which is over 4 standard deviations away from the expected length.

\clearpage
Our proposed solution is to use spherical linear interpolation \textit{slerp} instead of linear interpolation. We use a formula introduced by (Shoemake 85) in the context of great arc inbetweening for rotation animations:

\begin{figure*}[h]
  \centering
  \includegraphics[width=7cm]{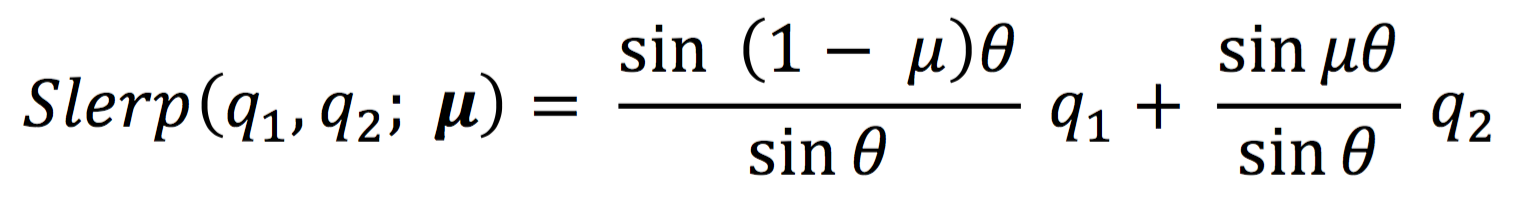}
\end{figure*}

This treats the interpolation as a great circle path on an n-dimensional hypersphere (with elevation changes). This technique has shown promising results on both VAE and GAN generative models and with both uniform and Gaussian priors (Figure 2).

\begin{figure}[h]
  \centering
  \includegraphics[width=8cm]{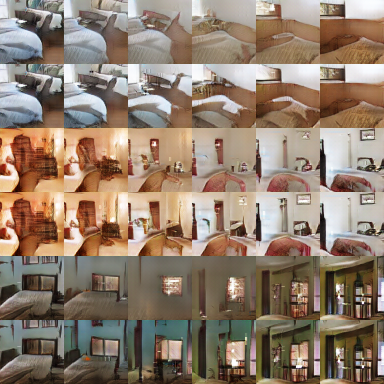}
  \caption{DCGAN (Radford 15) interpolation pairs with identical endpoints and uniform prior. In each pair, the top series is linear interpolation and the bottom is spherical. Note the weak generations produced by linear interpolation at the center, which are not present in spherical interpolation.}
\end{figure}

\subsection{Analogy}

Analogy has been shown to capture regularities in continuous space models. In the latent space of some linguistic models “King – Man + Woman” results in a vector very close to “Queen” (Mikolov et al., 2013). This technique has also been used in the context of deep generative models to solve visual analogies (Reed et al., 2015).

Analogies are usually written in the form:

\[ A : B :: C :  ? \]

Such a formation answers the question “What is the result of applying the transformation A:B to C?” In a vector space the solution generally proposed is to solve the analogy using vector math:

\[ (B - A) = (? - C) \]

\[ ? = C + B - A \]

Note that an interesting property of this solution is an implied symmetry:

\[ A : C :: B : ? \]

Because the same terms can be rearranged:

\[ (C - A) = (? - B) \]

\[ ? = B + C - A \]

Generative models that include an encoder for computing latent vectors given new samples allow for visual analogies. We have devised a visual representation for depicting analogies of visual generative networks called a “J-Diagram”. The J- Diagram uses interpolation across two dimensions two expose the manifold of the analogy. It also makes the symmetric nature of the analogy clear (Figure 3).

The J-Diagram also serves as a reference visualization across different model settings because it is deterministically generated from images which can be held constant. This makes it a useful tool for comparing results across epochs during training, after adjusting hyperparameters, or even across completely different model types (Figure 4).

\begin{figure}[h]
  \centering
  \includegraphics[width=8cm]{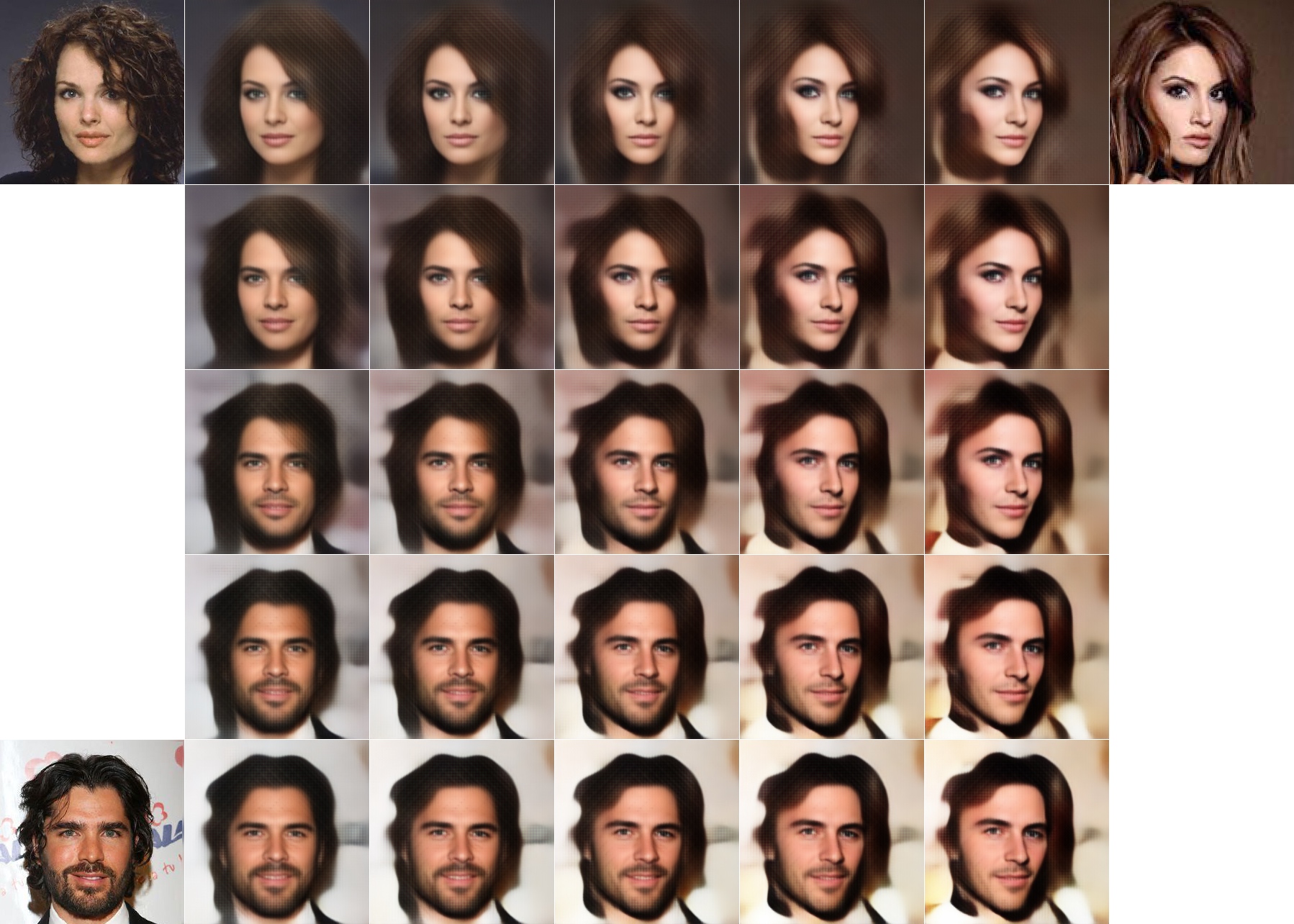}
  \caption{J-Diagram. The three corner images are inputs to the system, with the top left being the “source” (A) and the other two being “analogy targets” (B and C). Adjacent to each is the reconstruction resulting from running the image through both the encoder and decoder of the model. The bottom right image shows the result of applying the analogy operation (B + C) – A. All other images are interpolations using the slerp operator. (model: VAE from Lamb 16 on CelebA)}
\end{figure}

\begin{figure}[h]
  \centering
  \includegraphics[width=8cm]{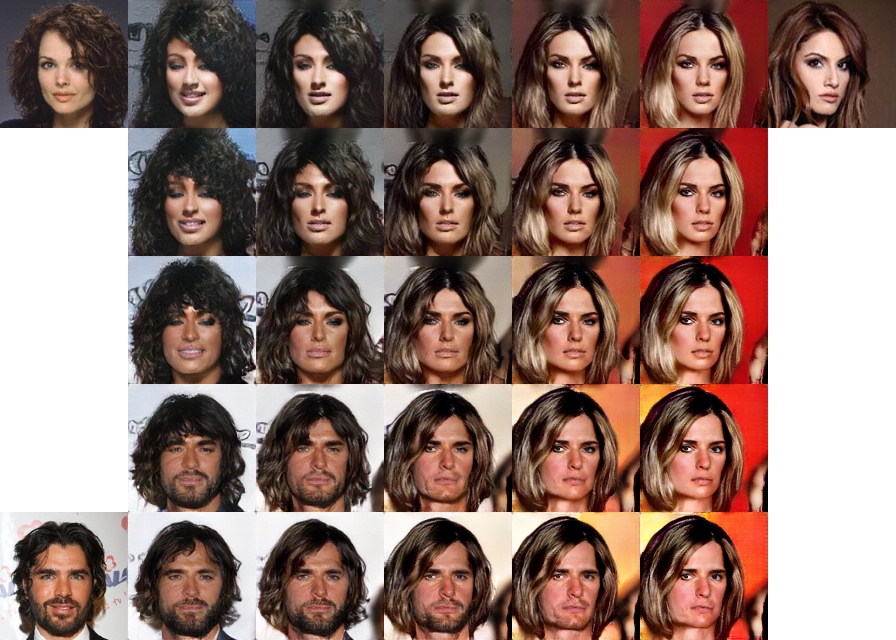}
  \caption{Same J-Diagram repeated with different model type. To facilitate comparisons (and demonstrate results are not cherry-picked) inputs selected are the first 3 images of the validation set. (model: GAN from Dumoulin 16 on CelebA)}
\end{figure}

\subsection{Manifold Traversal}

Generative models can produce a latent space that is not tightly packed, and the dimensionality of the latent space is often set artificially high. As a result, the manifold of trained examples is can be a subset of the latent space after training, resulting in dead zones in the expected prior.

If the model includes an encoder, one simple way to stay on the manifold is by only using out of sample encodings (ie: any data not used in training) in the latent space. This is a useful diagnostic, but is overly restrictive in a creative application context since it prevents the model from suggesting new and novel samples from the model. However, we can recover this ability by also including the results of operations on these encoded samples that stay close to the manifold, such as interpolation, extrapolation, and analogy generation.

Ideally, there would be a mechanism to discover this manifold within the latent space. In generative models with an encoder and ample out-of-sample data, we can instead precompute locations on the manifold with sufficient density, and later query for nearby points in the latent space from this known set. This offers a navigation mechanism based on hopping to nearest neighbors across a large database of encoded samples. When combined with interpolation, we call this visualization a Manifold Interpolated Neighbor Embedding
(MINE). A MINE grid is useful to visualize local patches of the latent space (Figure 5).

\begin{figure}[h]
    \centering
    \begin{subfigure}[b]{0.75\textwidth}
       \includegraphics[width=1\linewidth]{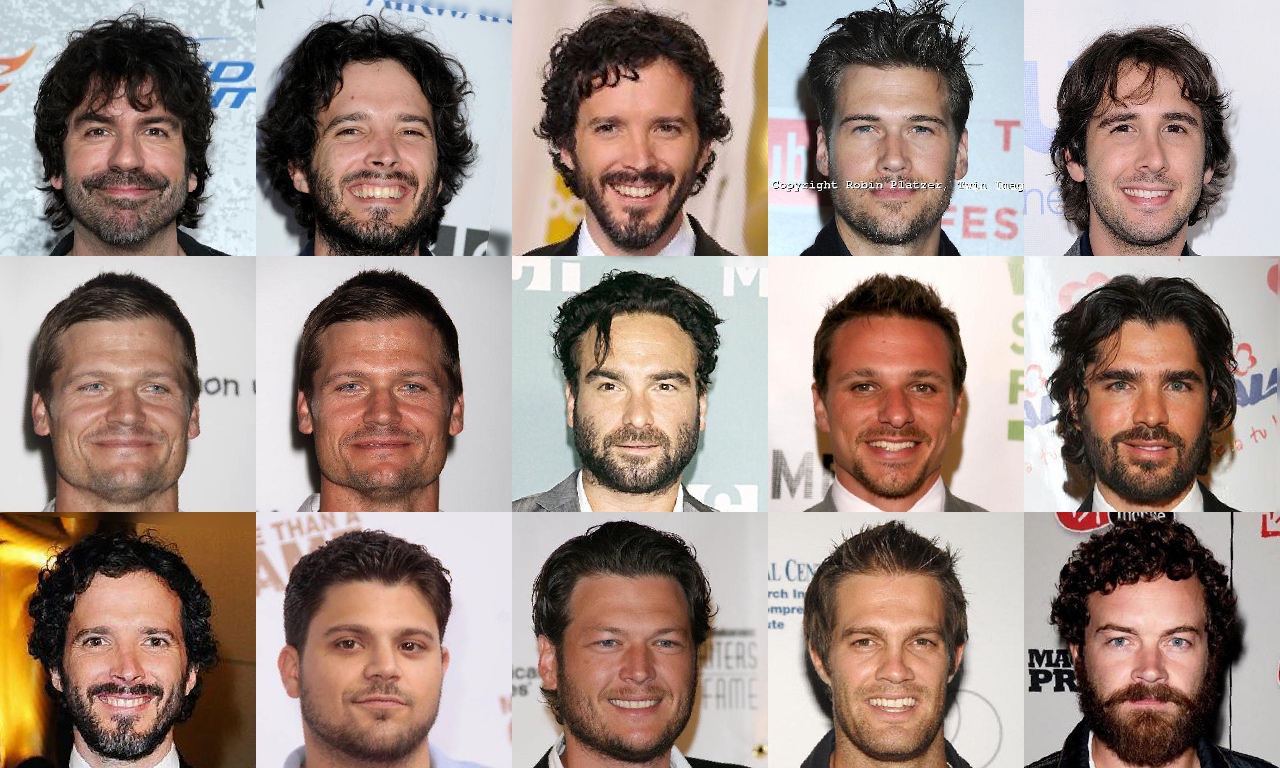}
       \caption{Nearest Neighbors are found and embedded into a two dimensional grid.}
       \label{fig:Ng1} 
    \end{subfigure}

    \begin{subfigure}[b]{0.75\textwidth}
       \includegraphics[width=1\linewidth]{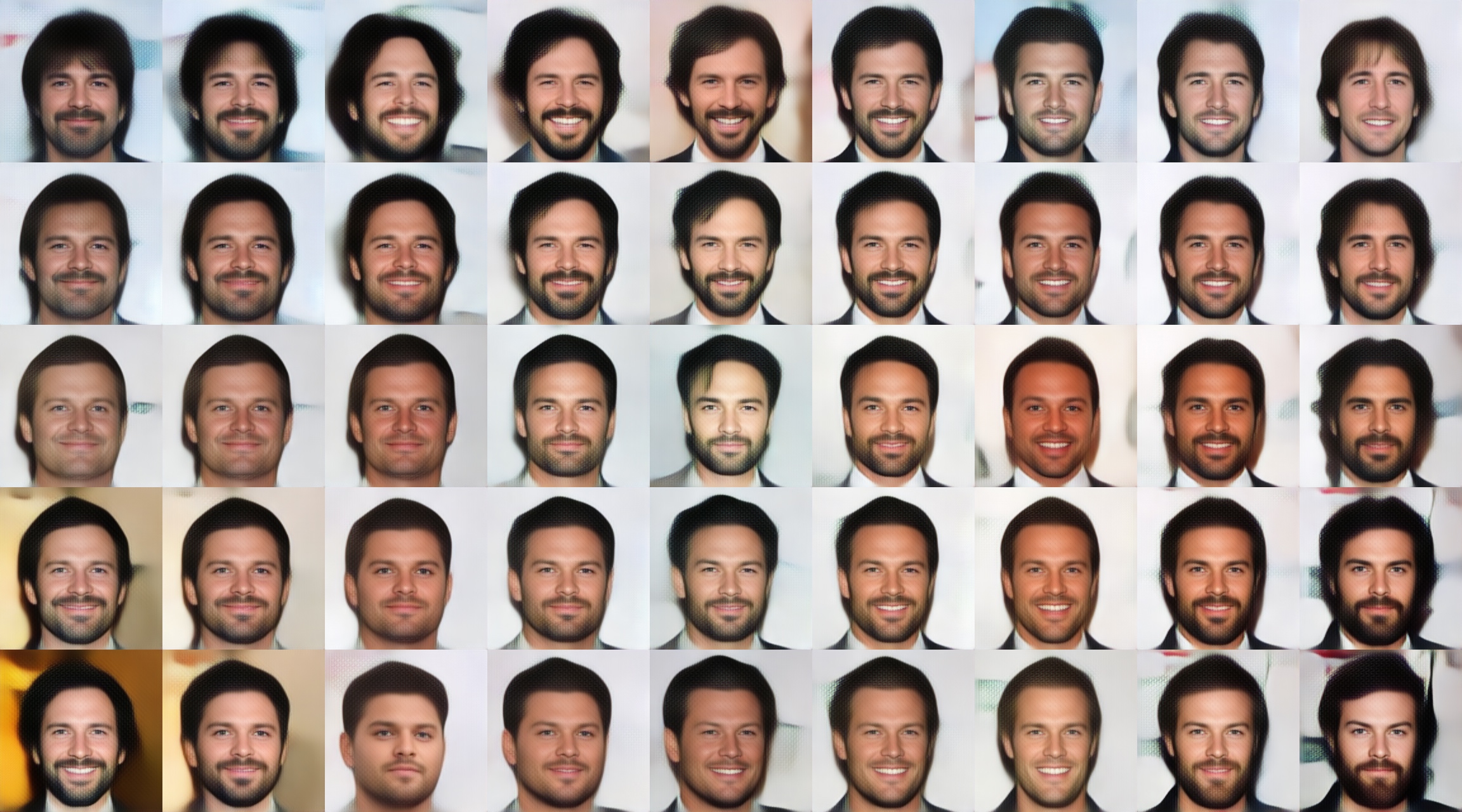}
       \caption{Reconstructions are spread with interpolation to expose areas between them.}
       \label{fig:Ng2}
    \end{subfigure}
    \caption{\label{fig:mine_grid} Example of local VAE manifold built using the 30k CelebA validation and test images as a dataset of out of sample features. The resulting MINE grid represents a small contiguous manifold of the larger latent space. (model: VAE from Lamb 16 on CelebA)}
\end{figure}

\section{Attribute Vectors}

Many generative models result in a latent space that is highly structured, even on purely unsupervised datasets (Radford et al., 2015). When combined with labeled data, attribute vectors can be computed using simple arithmetic. For example, a vector can be computed which represents the smile attribute, which by shorthand we call a smile vector. Following (Larsen et al., 2016), the smile vector can be computed by simply subtracting the mean vector for images without the smile attribute from the mean vector for images with the smile attribute. This smile vector can then be applied to in a positive or negative direction to manipulate this visual attribute on samples taken from latent space (Figure 6).

\begin{figure}[h]
  \centering
  \includegraphics[width=12cm]{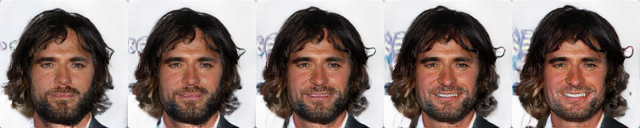}
  \caption{Traversals along the smile vector. (model: GAN from Dumoulin 16 on CelebA)}
\end{figure}

\subsection{Correlated Labels}

The approach of building attribute vectors from means of labeled data has been noted to suffer from correlated labels (Larsen et al., 2016). While many correlations would be expected from ground truths (eg: heavy makeup and wearing lipstick) we discovered others that appear to be from sampling bias. For example, male and smiling attributes have unexpected negative correlations because women in the CelebA dataset are much more likely to be smiling than men. (Table 1).

\begin{table}[h]
  \centering
  \begin{tabular}{ l | r | r | r }
    \cline{2-4}
            & Male & Not Male & \multicolumn{1}{|r|}{Total} \\
    \hline
    \multicolumn{1}{|l|}{Smiling} & 17\% & 31\% & \multicolumn{1}{|r|}{48\%} \\
    \hline
    \multicolumn{1}{|l|}{Not Smiling} & 25\% & 27\% & \multicolumn{1}{|r|}{52\%} \\
    \hline
    \multicolumn{1}{|l|}{Total} & 42\% & 58\% &  \\
    \cline{1-3}
  \end{tabular}
  \caption {
  Breakdown of CelebA smile versus male attributes. In the total population the smile attribute is almost balanced (48\% smile). But separating the data further shows that those with the male attribute smile only 42\% of the time while those without it smile 58\% of the time.}
\end{table}

\begin{figure}[h]
  \centering
  \includegraphics[width=8cm]{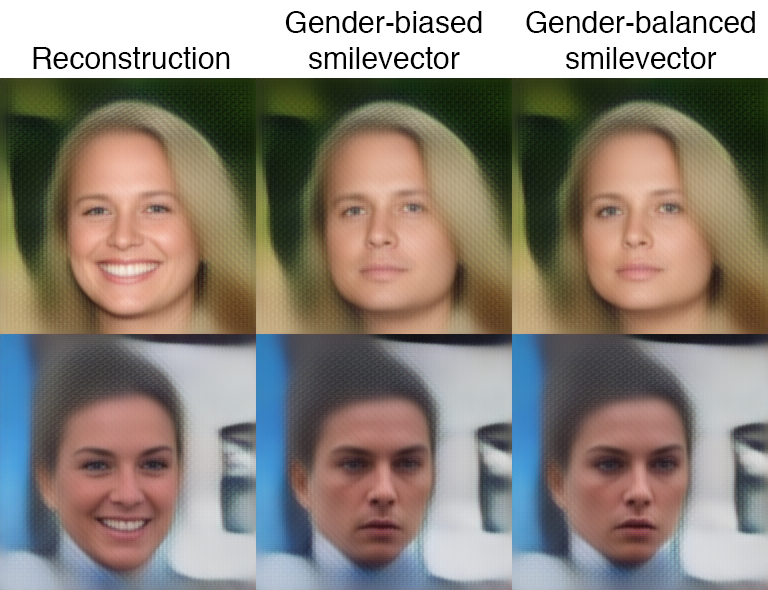}
  \caption{Initial attempts to build a smile vector suffered from sampling bias. The effect was that removing smiles from reconstructions (left) also added male attributes (center). By using replication to balance the data across both attributes before computing the attribute vectors, the gender bias was removed (right). (model: VAE from Lamb 16 on CelebA)}
\end{figure}

In an online service we setup to automatically add and remove smiles from images\footnote{\url{https://twitter.com/smilevector}}, we discovered this gender bias was visually evident in the results. Our solution was to use replication on the training data such that the dataset was balanced across attributes. This was effective because ultimately the vectors are simply summed together when computing the attribute vector (Figure 7).

This balancing technique can also be applied to attributes correlated due to ground truths. Decoupling attributes allows individual effects to be applied separately. As an example, the two attributes smiling and mouth open are highly correlated in the CelebA training set (Table 2). This is not surprising, as physically most people photographed smiling would also have their mouth open. However by forcing these attributes to be balanced, we can construct two decoupled attribute vectors. This allows for more flexibility in applying each attribute separately to varying degrees (Figure 8).

\begin{table}[h]
  \centering
  \begin{tabular}{ l | r | r | r }
    \cline{2-4}
            & Open Mouth & Not Open Mouth & \multicolumn{1}{|r|}{Total} \\
    \hline
    \multicolumn{1}{|l|}{Smiling} & 36\% & 12\% & \multicolumn{1}{|r|}{48\%} \\
    \hline
    \multicolumn{1}{|l|}{Not Smiling} & 12\% & 40\% & \multicolumn{1}{|r|}{52\%} \\
    \hline
    \multicolumn{1}{|l|}{Total} & 48\% & 52\% &  \\
    \cline{1-3}
  \end{tabular}
  \caption {
  CelebA smile versus open mouth attributes shows a strong symmetric correlation (greater than 3 to 1).}
\end{table}

\begin{figure}[h]
  \centering
  \includegraphics[width=12cm]{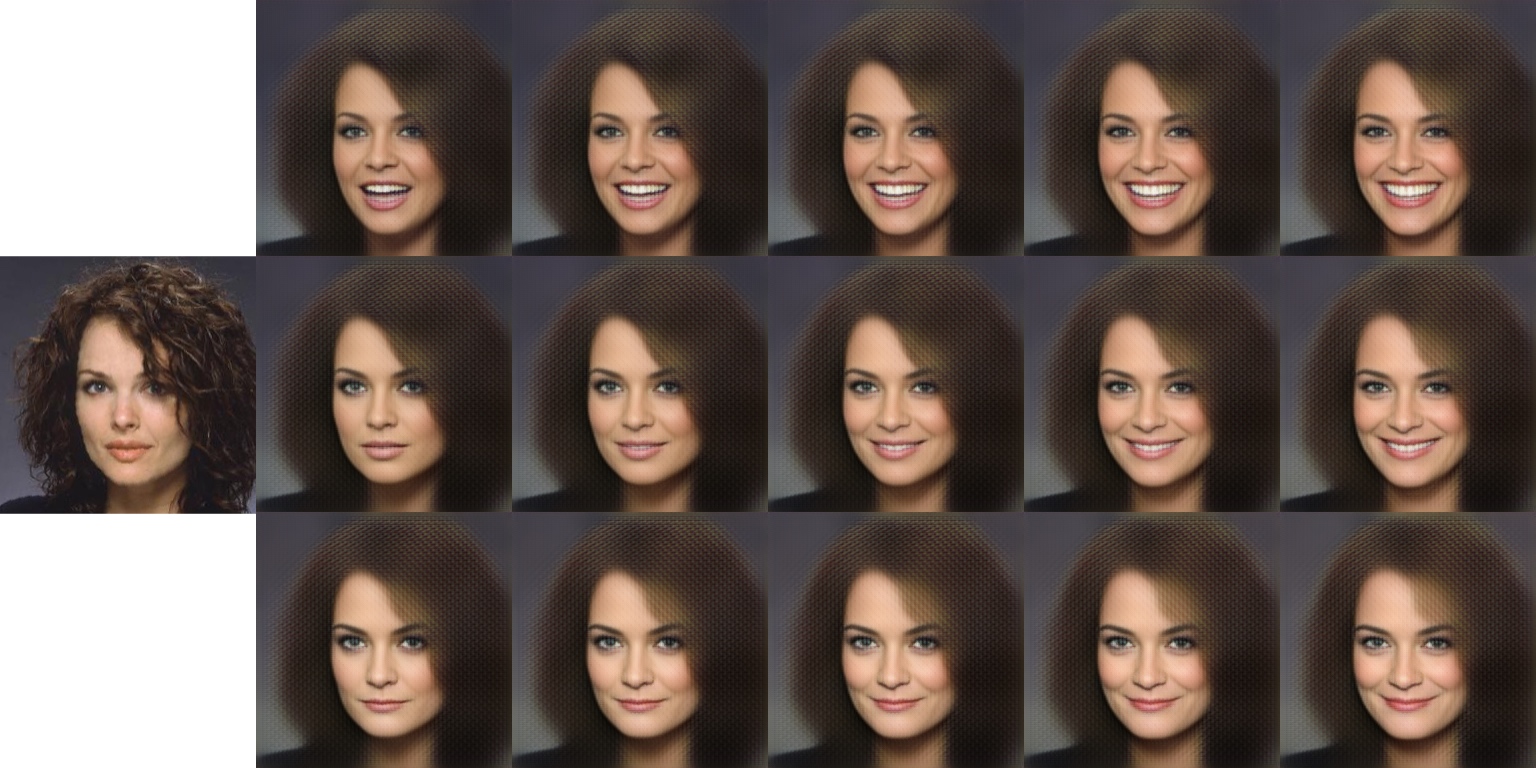}
  \caption{Decoupling attribute vectors for smiling (x-axis) and mouth open (y-axis) allows for more flexible latent space transformations. Input shown at left with reconstruction adjacent. (model: VAE from Lamb 16 on CelebA)}
\end{figure}

\subsection{Synthetic Attributes}

It has been noted that samples drawn from VAE based models is that they tend to be blurry (Goodfellow et al., 2014; Larsen et al., 2015). A possible solution to this would be to discover an attribute vector for “unblur”, and then apply this as a constant offset to latent vectors before decoding. CelebA includes a blur label for each image, and so a blur attribute vector was computed and then extrapolated in the negative direction. This was found to noticeably reduce blur, but also resulted in a number of unwanted artifacts such as increased image brightness.

We concluded this to be the result of human bias in labeling. Labelers appear more likely to label darker images as blurry, so this unblur vector was found to suffer from attribute correlation that also “lightened” the reconstruction. This bias could not be easily corrected because CelebA does not include a brightness label for rebalancing the data.

\begin{figure}[h]
  \centering
  \includegraphics[width=12cm]{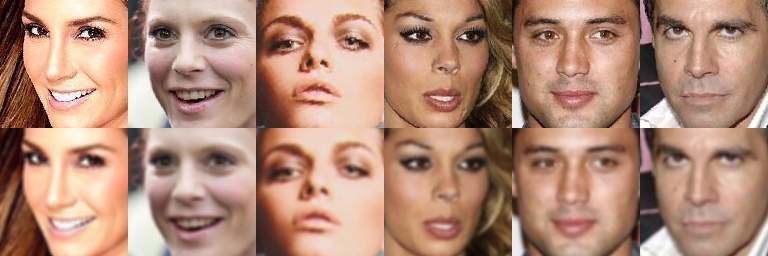}
  \caption{Zoomed detail from training images (top) and computed blurred images (bottom) were both encoded in order to determine a non-biased blur attribute vector.}
\end{figure}

For the blurring attribute, an algorithmic solution is available. We take a large set of images from the training set and process them through a Gaussian blur filter (figure 9). Then we run both the original image set and the blurred image set through the encoder and subtract the means of each set to compute a new attribute vector for blur. We call this a synthetic attribute vector because this label is derived from algorithmic data augmentation of the training set. This technique removes the labeler bias, is straightforward to implement, and resulted in samples closely resembling the reconstructions with less noticeable blur (figure 10).

\begin{figure}[h]
  \centering
  \includegraphics[width=12cm]{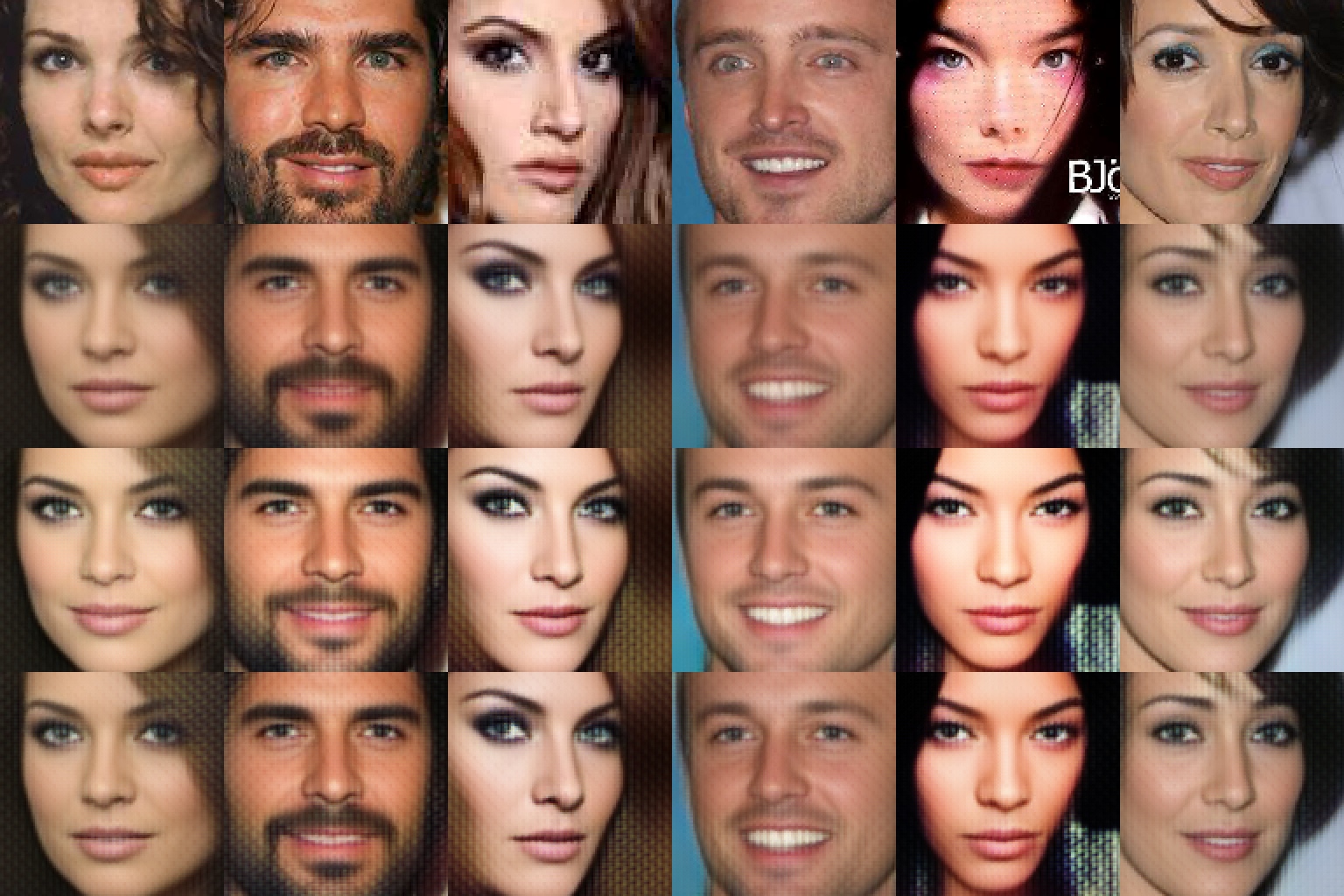}
  \caption{Zoomed detail of images from the validation set (top) are reconstructed (second row). Applying an offset in latent space based on the CelebA blur attribute (third row) does reduce noticeable blur from the reconstructions, but introduces visual artifacts including brightening due to attribute correlation. Applying an attribute vector instead computed from synthetic blur (bottom) yields images noticeably deblurred from the reconstructions and without unrelated artifacts.}
\end{figure}

\subsection{Quantitative Analysis with Classicication}

The results of applying attribute vectors visually to images are often quite striking. But it would be useful to be able to evaluate the relative efficacy of various attribute vectors to each other or across different models and hyperparamaters. Attribute vectors have potential to be widely applicable outside of images in domain specific latent spaces, and this provides additional challenges for quantifying their performance. For example, recent work has shown the ability to use latent space model as a continuous representation of molecules (Gómez 16); though it is straightforward to generate vectors for attributes such as solubility in these spaces, the evaluation of these operations to existing molecules would appear to require specific domain knowledge.
\begin{center}
\begin{table}[h]
 \begin{tabular}{||c c c c c c c c c||} 
 \hline
 Approach / Attribute & [1] & [2]
 W & [2]
 L & [3]
 ANet & [4]
 LNet & [4]
 Full & VAE AtDot & GAN AtDot \\ [0.5ex] 
 \hline\hline
 5 O.C. Shadow & 85 & 82 & 88 & 86 & 88 & \textbf{91} & 88 & 89 \\
 \hline
  Arched Eyebrow & 76 & 73 & 78 & 75 & 74 & \textbf{79} & 75 & 75 \\
 \hline 
 Attractive & 78 & 77 & \textbf{81} & 79 & 77 & \textbf{81} & 71 & 69 \\
 \hline 
 Bags Under Eye & 76 & 71 & 79 & 77 & 73 & 79 & \textbf{80} & 79 \\
 \hline 
 Bald & 89 & 92 & 96 & 92 & 95 & \textbf{98} & \textbf{98} & \textbf{98} \\
 \hline 
 Bangs & 88 & 89 & 92 & 94 & 92 & \textbf{95} & 90 & 90 \\
 \hline 
 Big Lips & 64 & 61 & 67 & 63 & 66 & 68 & \textbf{85} & 76 \\
 \hline 
 Big Nose & 74 & 70 & 75 & 74 & 75 & \textbf{78} & 77 & 77 \\
 \hline 
 Black Hair & 70 & 74 & 85 & 77 & 84 & \textbf{88} & 83 & 81 \\
 \hline 
 Blond Hair & 80 & 81 & 93 & 86 & 91 & \textbf{95} & 87 & 90 \\
 \hline 
 Blurry & 81 & 77 & 86 & 83 & 80 & 84 & \textbf{95} & \textbf{95} \\
 \hline 
 Brown Hair & 60 & 69 & 77 & 74 & 78 & 80 & 76 & \textbf{81} \\
 \hline 
 Bushy Eyebrow & 80 & 76 & 86 & 80 & 85 & \textbf{90} & 87 & 86 \\
 \hline 
 Chubby & 86 & 82 & 86 & 86 & 86 & 91 & \textbf{94} & \textbf{94} \\
 \hline 
 Double Chin & 88 & 85 & 88 & 90 & 88 & 92 & \textbf{95} & \textbf{95} \\
 \hline 
 Eyeglasses & 98 & 94 & 98 & 96 & 96 & \textbf{99} & 95 & 94 \\
 \hline 
 Goatee & 93 & 86 & 93 & 92 & 92 & \textbf{95} & 93 & 94 \\
 \hline 
 Gray Hair & 90 & 88 & 94 & 93 & 93 & \textbf{97} & 95 & 96 \\
 \hline 
 Heavy Makeup & 85 & 84 & \textbf{90} & 87 & 85 & \textbf{90} & 76 & 75 \\
 \hline 
 High Cheekbone & 84 & 80 & 86 & 85 & 84 & \textbf{87} & 81 & 68 \\
 \hline 
 Male & 91 & 93 & 97 & 95 & 94 & \textbf{98} & 81 & 80 \\
 \hline 
 Mouth Open & 87 & 82 & 93 & 85 & 86 & \textbf{92} & 78 & 67 \\
 \hline 
 Mustache & 91 & 83 & 93 & 87 & 91 & 95 & 95 & \textbf{96} \\
 \hline 
 Narrow Eyes & 82 & 79 & 84 & 83 & 77 & 81 & \textbf{93} & 89 \\
 \hline 
 No Beard & 90 & 87 & 93 & 91 & 92 & \textbf{95} & 85 & 84 \\
 \hline 
 Oval Face & 64 & 62 & 65 & 65 & 63 & 66 & \textbf{72} & 71 \\
 \hline 
 Pale Skin & 83 & 84 & 91 & 89 & 87 & 91 & \textbf{96} & \textbf{96} \\
 \hline 
 Pointy Nose & 68 & 65 & 71 & 67 & 70 & \textbf{72} & \textbf{72} & \textbf{72} \\
 \hline 
 Recede Hair & 76 & 82 & 85 & 84 & 85 & 89 & \textbf{93} & 92 \\
 \hline 
 Rosy Cheeks & 84 & 81 & 87 & 85 & 87 & 90 & \textbf{93} & \textbf{93} \\
 \hline 
 Sideburns & 94 & 90 & 93 & 94 & 91 & \textbf{96} & 94 & 94 \\
 \hline 
 Smiling & 89 & 89 & \textbf{92} & \textbf{92} & 88 & \textbf{92} & 87 & 68 \\
 \hline 
 Straight Hair & 63 & 67 & 69 & 70 & 69 & 73 & \textbf{80} & 79 \\
 \hline 
 Wavy Hair & 73 & 76 & 77 & 79 & 75 & \textbf{80} & 75 & 75 \\
 \hline 
 Earring & 73 & 72 & 78 & 77 & 78 & \textbf{82} & 81 & 81 \\
 \hline 
 Hat & 89 & 91 & 96 & 93 & 96 & \textbf{99} & 96 & 97 \\
 \hline 
 Lipstick & 89 & 88 & 93 & 91 & 90 & \textbf{93} & 79 & 77 \\
 \hline 
 Necklace & 68 & 67 & 67 & 70 & 68 & 71 & \textbf{88} & \textbf{88} \\
 \hline 
 Necktie & 86 & 88 & 91 & 90 & 86 & \textbf{93} & \textbf{93} & 92 \\
 \hline 
 Young & 80 & 77 & 84 & 81 & 83 & \textbf{87} & 78 & 81 \\
 \hline\hline 
 Average & 81 & 79 & 85 & 83 & 83 & \textbf{87} & 86 & 84 \\
 \hline
 \end{tabular}
\caption{Average Classification Accuracy on the CelebA dataset. Models for both VAE AtDot and GAN AtDot use unsupervised training with attribute vectors computed on the training set data after model training was completed. VAE AtDot from Lamb 16 with discriminative regularization disabled and GAN AtVec from Dumoulin 16. Other results are [1] Kumar 08, [2] Zhang 14, [3] Li 13, and [4] LeCun 94 as reported in Ehrlich 16.}
\end{table}
\end{center}

\clearpage

\begin{figure}[h]
    \centering
    \begin{subfigure}[t]{0.32\textwidth}
        \centering
        \includegraphics[width=4cm]{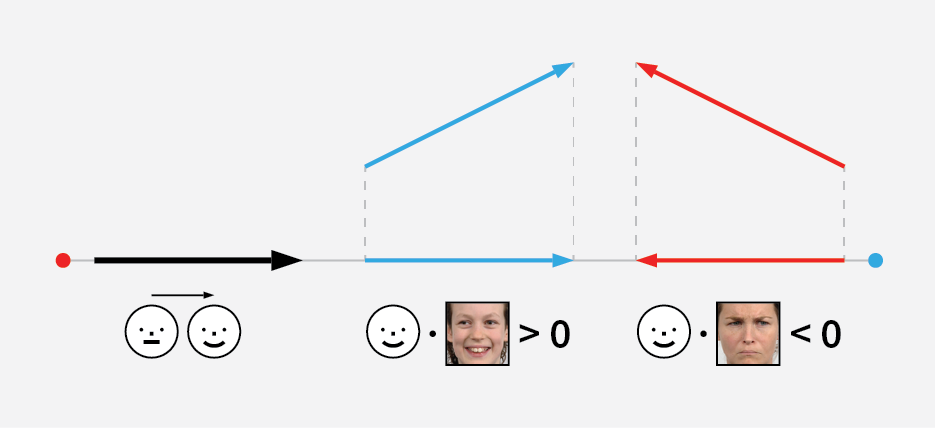}
        \caption{\label{fig:attribute_dotprod}Dot product.}
    \end{subfigure}
    \hfill
    \begin{subfigure}[t]{0.32\textwidth}
        \centering
        \includegraphics[width=4cm]{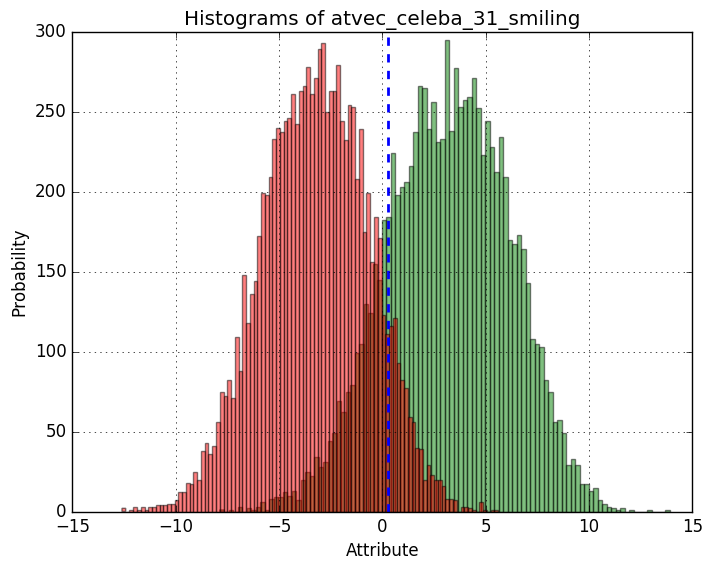}
        \caption{\label{fig:smile_histo}Histogram.}
    \end{subfigure}
    \begin{subfigure}[t]{0.32\textwidth}
        \centering
        \includegraphics[width=4cm]{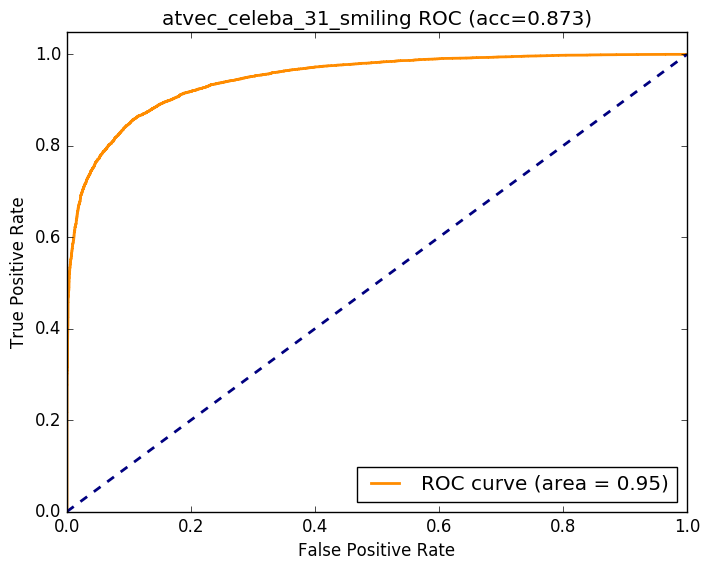}
        \caption{\label{fig:smile_roc}ROC curve.}
    \end{subfigure}
    \caption{\label{fig:j_diagrams}Example of using an attribute vector for classification. A smile vector is first constructed in latent space from labeled training data. By taking the dot product of this smile vector with the encoded representation of any face, a scalar smile score is generated (a). This score can be the basis for a binary classifier. The histogram shows the distribution of positive (green) and negative (red) scores on the CelebA validation set (b). The ROC curve for a smile vector based binary classifier is also shown (c).}
\end{figure}

We have found that very effective classifiers can be built in latent space using the dot product of an encoded vector with a computed attribute vector (figure 11A, 11B). We have named this technique for building classifiers on latent spaces AtDot, and models trained on unsupervised data have been shown to produce strong results on CelebA across most attributes (table 3). Importantly, these binary classifiers provide a quantitative basis to evaluate attribute vectors on a related surrogate task - their ability to be the basis for a simple linear classifier. This task is one of the most established paradigms in machine learning and provides a set of established tools such as ROC curves for offering new insights into the behavior of the attribute vectors across different hyperparameters or different models (figure 11C).

\section{Future Work}
Software to support most techniques presented in this paper is included in a python software library that can be used with various generative models\footnote{\url{https://github.com/dribnet/plat}}. We hope to continue to improve the library so that the techniques are applicable across a broad range of generative models. Attribute vector based classifiers also offer a promising way to evaluate the suitability of attribute vectors in domains outside of images.

We are investigating constructing a specially constructed prior on the latent space such that interpolations could be linear. This would simplify many of the latent space operations and might enable new types of operations.

Given sufficient test data, the extent to which an encoded dataset deviates from the expected prior should be quantifiable. Developing such a metric would be useful in understanding the structure of the different latent spaces including probability that random samples fall outside of the expected manifold of encoded data.

\subsubsection*{Acknowledgments}

I am thankful for the constructive feedback from readers including Ehud Ben-Reuven, Zachary Lipton, and Alex Champandard. I thank Victoria University of Wellington School of Design for supporting research on Creative Intelligence. I also thank the vibrant machine learning and creative coding communities on twitter for their support and encouragement.

\subsubsection*{References}

\small

Dumoulin, Vincent, Belghazi, Ishmael, Poole, Ben, Lamb, Alex, Arjovsky, Martin, Mastropietro, Olivier Courville, Aaron. Adversarially Learned Inference. {\url{https://arxiv.org/abs/1606.00704}} 2016

Ehrlich, Max, Shields, Timothy J., Almaev, Timur, Amer, Mohamed R. Facial Attributes Classification Using Multi-Task Representation Learning. The IEEE Conference on Computer Vision and Pattern Recognition (CVPR) Workshops. 2016.

Goodfellow, Ian, Pouget-Abadie, Jean, Mirza, Mehdi, Xu, Bing, Warde- Farley, David, Ozair, Sherjil, Courville, Aaron, and Bengio, Yoshua. Generative adversarial nets. In Ghahramani, Z., Welling, M., Cortes, C., Lawrence, N.D., and Weinberger, K.Q. (eds.), Advances in Neural Information Processing Systems 27, pp. 2672–2680. Curran Associates, Inc., 2014.

Gómez-Bombarelli, Rafael, Duvenaud, David, Miguel, Hernández-Lobato, José, Aguilera-Iparraguirre, Jorge, Hirzel, Timothy D., Adams, Ryan P., and Alán Aspuru-Guzik. Automatic chemical design using a data-driven continuous representation of molecules. {\url{https://arxiv.org/abs/1610.02415}} 2016

Kingma, Diederik P. and Welling, Max. Auto-encoding variational Bayes. In Proceedings of the International Conference on Learning Representations, 2014.

Kumar, N., Belhumeur, P., and Nayar, S. Facetracer: A search engine for large collections of images with faces. In ECCV, 2008.

Lamb, Alex, Dumoulin, Vincent, Courville, Aaron. Discriminative Regularization for Generative Models. {\url{http://arxiv.org/abs/1602.03220}} 2016

Larsen, Anders Boesen Lindbo, Sønderby, Søren Kaae, Larochelle, Hugo, Winther, Ole. Autoencoding beyond pixels using a learned similarity metric. {\url{https://arxiv.org/abs/1512.09300}} 2016

LeCun, Y. and Bengio., Y. Word-level training of a handwritten word recognizer based on convolutional neural networks. In ICPR, 1994.

Li, J., and Zhang, Y. Learning surf cascade for fast and accurate object detection. In CVPR, 2013.

Makhzani, Alireza, Shlens, Jonathon, Jaitly, Navdeep, Goodfellow, Ian, Brendan, Frey. Adversarial Autoencoders. {\url{http://arxiv.org/abs/1511.05644}. 2016.

Mikolov, Tomas, Yih, Scott Wen-tau, Zweig, Geoffrey. Linguistic Regularities in Continuous Space Word Representations. NAACL-HLT, 2013.

Radford , Alec, Metz , Luke, Chintala, Soumith. Unsupervised Representation Learning with Deep Convolutional Generative Adversarial Networks. Advances in Neural Information Processing Systems, 2015.

Reed, Scott E, Zhang, Yi, Zhang, Yuting, and Lee, Honglak. Deep visual analogy-making. In Cortes, C., Lawrence, N.D., Lee, D.D., Sugiyama, M., Garnett, R., and Garnett, R. (eds.), Advances in Neural Information Processing Systems 28, pp. 1252–1260. Curran Associates, Inc., 2015

Shoemake, Ken. Animating rotation with quaternion curves. In ACM Siggraph, 19(3):245–254, 1985.

Zhang, N., Paluri, M., Ranzato, M.,  Darrell, T., and Bourdev, L. Panda: Pose aligned networks for deep attribute modeling. In CVPR, 2014.

\end{document}